%% file: main.tex
\documentclass[10pt,twocolumn,letterpaper]{article}

\usepackage[pagenumbers]{cvpr} 

\input{preamble}

\definecolor{cvprblue}{rgb}{0.21,0.49,0.74}
\usepackage[pagebackref,breaklinks,colorlinks,allcolors=cvprblue]{hyperref}

\def\paperID{18596} 
\def\confName{CVPR}
\def\confYear{2026}

\title{NCSTR: Node-Centric Decoupled Spatio-Temporal Reasoning for Video-based Human Pose Estimation}

\author{Quang Dang Huynh\\
Griffith University\\
{\tt\small quangdang.huynh@griffithuni.edu.au}
\and
Xuefei Yin\\
Griffith University\\
{\tt\small x.yin@griffith.edu.au}
\and
Andrew Busch\\
Griffith University\\
{\tt\small a.busch@griffith.edu.au}
\and
Hugo G. Espinosa\\
Griffith University\\
{\tt\small h.espinosa@griffith.edu.au}
\and
Alan Wee-Chung Liew\\
Griffith University\\
{\tt\small a.liew@griffith.edu.au}
\and
Matthew Worsey\\
Griffith University\\
{\tt\small m.worsey@griffith.edu.au}
\and
Yanming Zhu\\
Griffith University\\
{\tt\small yanming.zhu@griffith.edu.au}
}

\begin{document}
\maketitle
\input{sec/0_abstract}    
\input{sec/1_intro}

\input{sec/2_method}
\input{sec/3_experiment}

{
    \small
    \bibliographystyle{ieeenat_fullname}
    \bibliography{main}
}


\end{document}

%% file: preamble.tex
\usepackage{amsmath,amssymb,amsfonts}
\usepackage{mathtools}
\usepackage{enumitem}
\usepackage{bm}
\usepackage{physics}
\usepackage{booktabs}
\usepackage{microtype}
\usepackage{subcaption}

\usepackage{microtype}



%% file: sec/0_abstract.tex
\begin{abstract}
Video-based human pose estimation remains challenged by motion blur, occlusion, and complex spatiotemporal dynamics. Existing methods often rely on heatmaps or implicit spatio-temporal feature aggregation, which limits joint topology expressiveness and weakens cross-frame consistency.
To address these problems, we propose a novel node-centric framework that explicitly integrates visual, temporal, and structural reasoning for accurate pose estimation. 
First, we design a visuo-temporal velocity-based joint embedding that fuses sub-pixel joint cues and inter-frame motion to build appearance- and motion-aware representations.
Then, we introduce an attention-driven pose-query encoder, which applies attention over joint-wise heatmaps and frame-wise features to map the joint representations into a pose-aware node space, generating image-conditioned joint-aware node embeddings.
Building upon these node embeddings, we propose a dual-branch decoupled spatio-temporal attention graph that models temporal propagation and spatial constraint reasoning in specialized local and global branches.
Finally, a node-space expert fusion module is proposed to adaptively fuse the complementary outputs from both branches, integrating local and global cues for final joint predictions.
Extensive experiments on three widely used video pose benchmarks demonstrate that our method outperforms state-of-the-art methods. The results highlight the value of explicit node-centric reasoning, offering a new perspective for advancing video-based human pose estimation.
\end{abstract}

%% file: sec/1_intro.tex
\section{Introduction}
\label{sec:intro}

Video-based human pose estimation (VHPE) is essential for understanding human actions and interactions in dynamic visual environments~\cite{xu2021vipnas, dang2022relation, yu2023bidirectionally, tan2024diffusionregpose}. It underpins numerous downstream applications such as motion analysis, sports analytics, human-computer interaction, and autonomous systems~\cite{zheng2022multi, ko2023pose, neupane2024survey}. Despite recent progress, accurate VHPE remains challenging due to motion blur, occlusion, large pose variation, and complex temporal dynamics~\cite{lan2022vision, zheng2023deep, yin2024survey, an2024sharpose}.

Prior works address temporal consistency through feature alignment or multi-frame heatmap refinement~\cite{liu2021deep, liu2022temporal}, yet they often rely on global propagation and lack explicit joint-level modeling. Generative methods such as diffusion-based and autoregressive temporal models~\cite{feng2023diffpose, nguyen2025autoregressive} improve long-range temporal reasoning but incur high computational overhead. Transformer-based methods~\cite{chen2025causal, jiao2025spatiotemporal, jiao2025optimizing} introduce keypoint tokens or adaptive masking to enhance robustness, although their reasoning often remains at a coarse global level with limited structural constraints. Recent graph or state space formulations~\cite{yang2021learning, wu2024joint, feng2025high, he2024video} try to model skeletal topology and motion, yet spatial and temporal cues are often entangled within unified modules, restricting fine-grained structural reasoning and joint-aware temporal propagation.

To address these limitations, we propose a node-centric decoupled spatio-temporal reasoning framework that explicitly integrates visual, temporal, and structural cues for accurate VHPE. First, a Visuo-Temporal Velocity-based Joint Embedding is proposed to encode sub-pixel joint estimates, inter-frame velocity,  and visibility to form expressive motion-aware joint representations. An attention-based Pose-Query Encoder further refines these representations and maps them into a joint-aware node space enriched with image-conditioned contextual information. Based on these expressive node embeddings, we design a Dual-branch decoupled Spatio-Temporal Attention Graph that performs temporal state propagation followed by spatial constraint reasoning in two specialized branches: one emphasizing local anatomical consistency and the other emphasizing global contextual coherence. Finally, a Node-Space Expert Fusion module is designed to adaptively fuse the outputs from both branches, integrating local and global cues to produce precise and stable joint predictions.
Extensive experiments on three video pose benchmarks demonstrate that our method surpasses state-of-the-art (SOTA) methods. These results highlight the importance of explicit node-level representation, fine-grained structural modeling, and decoupled spatio-temporal graph reasoning for advancing VHPE.

The key contributions are summarized as follows:
\begin{itemize}
    \item We propose a \textbf{Visuo-Temporal Velocity-based Joint Embedding (VTVJE)} that explicitly encodes joint position, inter-frame velocity, and visibility, producing expressive joint representations.
    \item We design an \textbf{attention-based Pose-Query Encoder (PQE)} that maps the VTVJE representations into an image-conditioned joint-aware node space.

    \item We develop a \textbf{Dual-branch decoupled Spatio-Temporal Attention Graph (DSTAG)} that separately models temporal propagation and spatial constraint reasoning in specialized local and global branches.
    \item We introduce a \textbf{Node-Space Expert Fusion (NSEF)} module that adaptively integrates local and global node experts for accurate and stable joint predictions.

    \item Our method achieves SOTA performance on video pose benchmarks, demonstrating the effectiveness of explicit node-centric and decoupled spatio-temporal reasoning.

\end{itemize}

\begin{figure*}[htbp]
\centering
\includegraphics[width=1\linewidth]{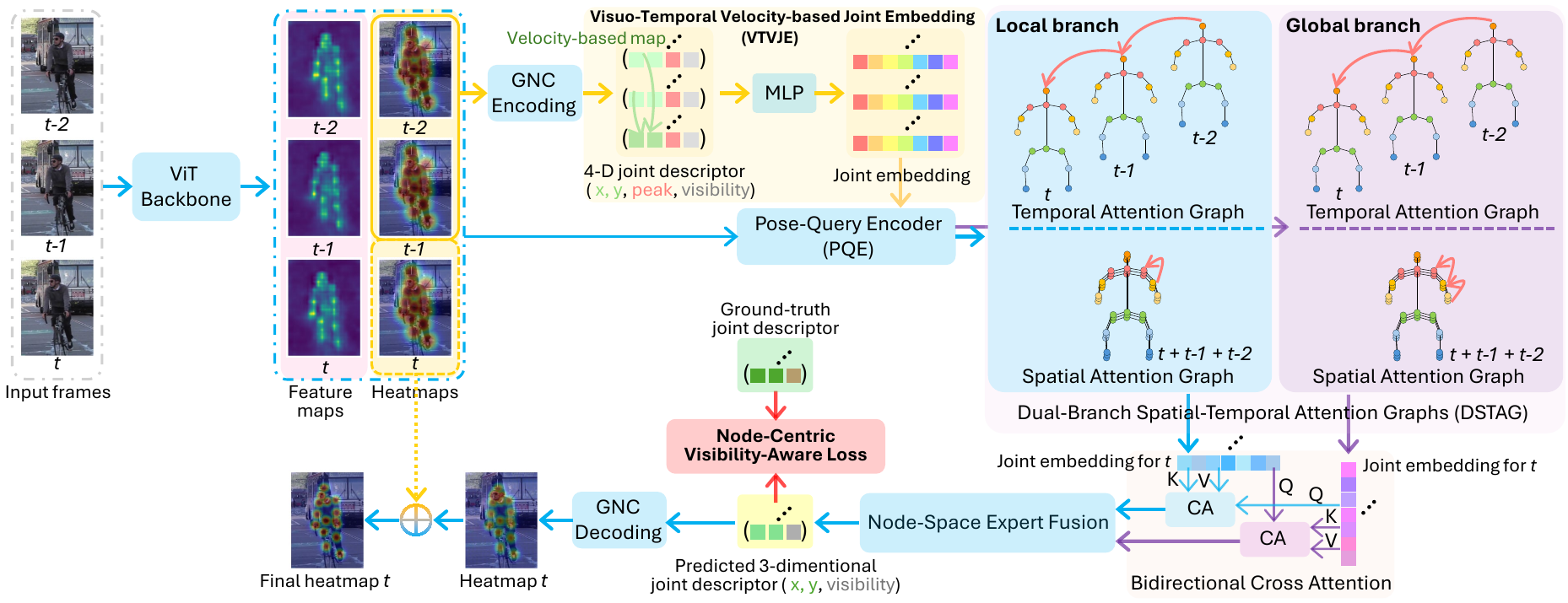}
\caption{
Overview of the proposed node-centric decoupled spatio-temporal attention graph for VHPE. Given three frames, initial pose estimates and backbone features are extracted.
A VTVJE encodes sub-pixel joint locations, velocity, peak score, and visibility into a joint descriptor, which the PQE refines into node embeddings.
The embeddings are processed by the DSTAG: a \textit{local} branch enforcing fine-grained structure and a \textit{global} branch capturing long-range context.
The NSEF module then fuses both branches to produce node predictions.
Finally, a fixed GNC-Decoding module renders heatmaps from node predictions, supervised by a visibility-aware node loss.
}

\label{fig:workflow}
\end{figure*}

\section{Related Work}
\label{sec:related_work}

\subsection{Video-Based Human Pose Estimation}
Early multi-frame methods extended image-based estimators by propagating temporal features or refining heatmaps across frames~\cite{liu2021deep, liu2022temporal}. Although they improve temporal smoothness, their reliance on global feature alignment limits cross-frame consistency and lacks explicit joint-level reasoning. Tracking-based pipelines~\cite{wang2020combining} integrate detection and association but struggle with fast motion and occlusion due to their dependence on bounding-box matching. More recent methods focus on explicit temporal modeling. Regression-based formulations~\cite{he2024video} decouple spatial and temporal aggregation to improve stability without heatmap decoding. Diffusion-driven models~\cite{feng2023diffpose, nguyen2025autoregressive} enhance long-range reasoning but incur significant computational overhead, limiting scalability. These challenges motivate more structured and efficient temporal reasoning frameworks.

\subsection{Attention-Based Temporal Modeling}
Transformer-based methods have shown strong temporal modeling capacity for VHPE. Temporal–interim pose synthesis~\cite{zhang2025temporal} enables attention over intermediate states but still captures temporal relations implicitly without structural constraints. Causal multitask learning~\cite{chen2025causal} improves robustness via keypoint tokens and causal masks, while spatiotemporal fusion models~\cite{jiao2025spatiotemporal, jiao2025optimizing} adapt attention to human and joint regions. Despite their progress, these methods rely on global attention layers and struggle to encode explicit skeletal structure or disentangle spatial and temporal dependencies.

\subsection{GNN-Based and Motion-Aware Modeling}
Graph-based methods offer structural priors for modeling joints and relations. Early GNN-based motion propagation~\cite{yang2021learning} modeled joint trajectories but lacked appearance and visibility cues. Joint-motion mutual learning~\cite{wu2024joint} adds optical flow with joint features but fuses spatial and temporal cues within one module, limiting fine-grained relational reasoning. Other works enhance temporal alignment via mutual information or decoupled regression~\cite{liu2022temporal, he2024video}, but they still rely on global features or heatmap propagation and do not provide explicit multimodal joint representations capturing sub-pixel cues, velocity, and confidence.

In contrast, our method introduces a node-centric, motion-aware formulation that unifies visual, temporal, and structural reasoning. By leveraging a visuo-temporal velocity-based joint embedding and a dual-branch decoupled spatiotemporal graph, we achieve explicit joint-level modeling and structurally disentangled reasoning.

%% file: sec/2_method.tex
\section{Method}

\subsection{Problem Formulation}
We follow a top-down setting for video human pose estimation. Let
$\{I_t\}_{t=1}^{T}$ denote the input video, where $I_t$ is the frame at time
$t$. A human detector is applied to each frame to obtain bounding boxes for all
persons~\cite{redmon2018yolov3}, following the exact detector protocol in~\cite{liu2021deep,he2024video,liu2022temporal}. Each box is enlarged by a factor of $1.25$,
cropped, and resized to $384\times288$.
For a person $i$ at time $t$, we define a causal temporal clip
$
\boldsymbol{\mathcal{P}}_{t}^{i}
=
\{P_{t-k}^{i},\dots,P_{t-1}^{i},P_{t}^{i}\},
$
where $P_{t}^{i}$ is the cropped person image.
We fix the window size to $k=2$:
\begin{equation}
\boldsymbol{\mathcal{P}}_{t}^{i}
=
\{P_{t-2}^{i},\,P_{t-1}^{i},\,P_{t}^{i}\}.
\end{equation}
Given this clip $\boldsymbol{\mathcal{P}}_{t}^{i}$, our goal is to
estimate the pose of $P_{t}^{i}$. 
Unlike most existing methods that require access to future frames, our method is fully online and strictly causal.

We introduce a node-centric geometric presentation. Let $J$ be the number of joints. For each joint $j \in \{1, \dots, J\}$, we define a \emph{geometric joint representation}
\begin{equation}
\mathbf{n}_{t,j}^{i}
=
\bigl[
x^{\mathrm{norm}}_{t,j},
y^{\mathrm{norm}}_{t,j},
v_{t,j}
\bigr]
\in [0,1]^3,
\end{equation}
where $x^{\mathrm{norm}}_{t,j}$ and $y^{\mathrm{norm}}_{t,j}$ are normalized coordinates, and $v_{t,j}$ is the joint visibility score. Stacking all joints gives
$\mathbf{N}_{t}^{i} \in [0,1]^{J \times 3}.$
Therefore, given the clip $\boldsymbol{\mathcal{P}}_{t}^{i}$, our method predicts $\mathbf{N}_{t}^{i}$ for person $i$ at frame $t$. 

Our framework learns directly in the joint (node) space using image-conditioned node embeddings, with the geometric representation as the final prediction target, as illustrated in \cref{fig:workflow}. The next sections describe its main components: Visuo-Temporal Velocity-based Joint Embedding (\cref{sec:VTVJE}), Pose-Query Encoder (\cref{sec:PQE}), Dual-branch decoupled Spatiotemporal Attention Graph (\cref{sec:DSTAG}), Node-Space Expert Fusion (\cref{sec:NSEF}), and Geometric Node Codec (\cref{sec:GNC}).

\textbf{Notation.}
Let $T$ denote the number of frames, $I$ the number of persons, $J$ the number
of joints, $H\times W$ the heatmap resolution, and $C$ the number of feature
channels. We index frames by $t$,
persons by $i$, and joints by $j$ throughout the paper.
Given a temporal window $\{I_{t-2}, I_{t-1}, I_{t}\}$, a ViT backbone
produces frame-wise features and initial joint-wise heatmaps. The heatmaps $\mathcal{H}\in\mathbb{R}^{T\times J\times H\times W}$ are generated by a standard prediction head and serve as joint priors. The features are extracted from the final transformer block and upsampled to the heatmap resolution, denoted by
$\boldsymbol{F}\in\mathbb{R}^{T\times C\times H\times W}$.

\subsection{Visuo-Temporal Velocity-based Joint Embedding (VTVJE)}
\label{sec:VTVJE}

The VTVJE encodes joint position, visibility, and inter-frame motion velocity to produce expressive joint embeddings.

\subsubsection{Past Joint Embeddings}
For each past frame $t' \in \{t-2, t-1\}$ with heatmap 
$\mathcal{H}_{t'} \in \mathbb{R}^{J \times H \times W}$, 
sub-pixel joint coordinates are obtained using a second-order Taylor expansion. 
Let $(h^{0}_{t',j}, w^{0}_{t',j})$ be the integer peak of joint $j$. On its $3 \times 3$ neighborhood, let $\nabla \mathcal{H}$ and $\mathbf{H}$ denote the gradient and Hessian. The refined sub-pixel coordinate is

\begin{equation}
\boldsymbol{p}^{i}_{t',j}
=
\boldsymbol{p}_{t',j}
-\mathbf{H}_{t',j}^{-1}\,\nabla \mathcal{H}_{t',j},
\end{equation}
where 
$\boldsymbol{p}_{t',j}=[w^{0}_{t',j},\, h^{0}_{t',j}]^\top$ is the integer peak
location. After normalization, we construct a 4-dimensional descriptor:

\begin{equation}
\mathbf{p}_{t',j}^{i}
=
\bigl[
x^{\mathrm{norm}}_{t',j},
y^{\mathrm{norm}}_{t',j},
v^{\mathrm{peak}}_{t',j},
v^{\mathrm{vis}}_{t',j}
\bigr]
\in [0,1]^4,
\end{equation}
where $v^{\mathrm{vis}}_{t',j}\in\{0,1\}$ is obtained by thresholding the peak value at $0.35$. A small MLP maps this descriptor to a $D$ dimensional embedding:

$\mathbf{q}_{t',j}^{i} = 
f_{\mathrm{MLP}}\left(\mathbf{p}_{t',j}^{i}\right)
\in \mathbb{R}^{D}.$

\subsubsection{Current Joint Embeddings}
\label{sec:current_joint_em}

For the current frame, we introduce a velocity-based descriptor.
Given the past sub-pixel coordinates $\boldsymbol{p}^{i}_{t-2,j}$ and $\boldsymbol{p}^{i}_{t-1,j}$, we 
estimate the displacement
$\Delta x_{t,j}=x^{i}_{t-1,j}-x^{i}_{t-2,j}$ and 
$\Delta y_{t,j}=y^{i}_{t-1,j}-y^{i}_{t-2,j}$. 
A linear extrapolation yields the motion estimate
$
\tilde{x}^{i}_{t,j}=x^{i}_{t-1,j}+\Delta x_{t,j} ~\text{and }~
\tilde{y}^{i}_{t,j}=y^{i}_{t-1,j}+\Delta y_{t,j}.
$

The normalized coordinates are concatenated with the peak value and last frame's visibility to form:
\begin{equation}
\mathbf{p}_{t,j}^{i}
=
\bigl[
x^{\mathrm{norm}}_{t,j},
y^{\mathrm{norm}}_{t,j},
v^{\mathrm{peak}}_{t-1,j},
v^{\mathrm{vis}}_{t-1,j}
\bigr]
\in [0,1]^4.
\end{equation}
This descriptor is also projected to a $D$-dimensional embedding via a 
lightweight MLP:

$\mathbf{q}_{t,j}^{i} = 
f_{\mathrm{MLP}}\left(\mathbf{p}_{t,j}^{i}\right)
\in \mathbb{R}^{D}.$

\subsection{Pose-Query Encoder (PQE)}
\label{sec:PQE}

The PQE enriches joint representations by applying joint-conditioned attention over frame features and heatmaps, yielding image-conditioned node embeddings robust to occlusion and noisy heatmap peaks.

Each joint provides a query vector, and every spatial location in the feature map and heatmap provides a key-value pair. For each frame $t' \in \{t-2,t-1,t\}$, given a joint query 
$\boldsymbol{q}_{t',j}\in\mathbb{R}^{D}$ and a feature map 
$\boldsymbol{F}_{t'}\in\mathbb{R}^{C\times H\times W}$, PQE applies shared linear projections $W_Q\in\mathbb{R}^{D\times C_q},
W_K,W_V\in\mathbb{R}^{C\times C_q},
W_O\in\mathbb{R}^{C_q\times F}$
to produce 
\begin{equation}
\left\{
\begin{aligned}
\boldsymbol{Q}_{t',j} &= \boldsymbol{q}_{t',j} W_Q \in \mathbb{R}^{C_q},\\
\boldsymbol{K}_{t',h,w} &= \boldsymbol{F}_{t',:,h,w}^{\top} W_K \in \mathbb{R}^{C_q},\\
\boldsymbol{V}_{t',h,w} &= \boldsymbol{F}_{t',:,h,w}^{\top} W_V \in \mathbb{R}^{C_q}.
\end{aligned}
\right.
\end{equation}

\paragraph{Attention with Priors and Masks.}

Joint-wise attention over the spatial grid is computed using a learnable temperature. The raw attention logit is
\begin{equation}
\ell_{t',j,h,w}
=
\frac{\langle \boldsymbol{Q}_{t',j},\boldsymbol{K}_{t',h,w}\rangle}
{\max(\tau,\varepsilon)}
-10^{4}(1-\mathbf{M}_{t',j,h,w})
\end{equation}
where $\tau$ is a learnable temperature, $\varepsilon=10^{-6}$ ensures numerical stability, and $\mathbf{M}$ is a binary mask.

When heatmaps are used as priors, we simply add
$\ell_{t',j,h,w}
= \ell_{t',j,h,w} + \mathcal{H}_{t',j,h,w}.$

\paragraph{Local and Global Masks.}
For the current frame $t$, we compute a motion-adaptive radius based on the 
estimated joint velocity:
\begin{equation}
r_{t,j}
=
\min\bigl(\max(v_{t,j},\,r_{\min}),\,r_{\max}\bigr),
\end{equation}
where 
\(v_{t,j}=\sqrt{(\Delta x_{t,j})^{2}+(\Delta y_{t,j})^{2}}\) (\cref{sec:current_joint_em}), \(r_{\min}=3\) prevents collapse, and \(r_{\max}=15\) limits spread.

\begin{figure}
    \centering
    \includegraphics[width=1\linewidth]{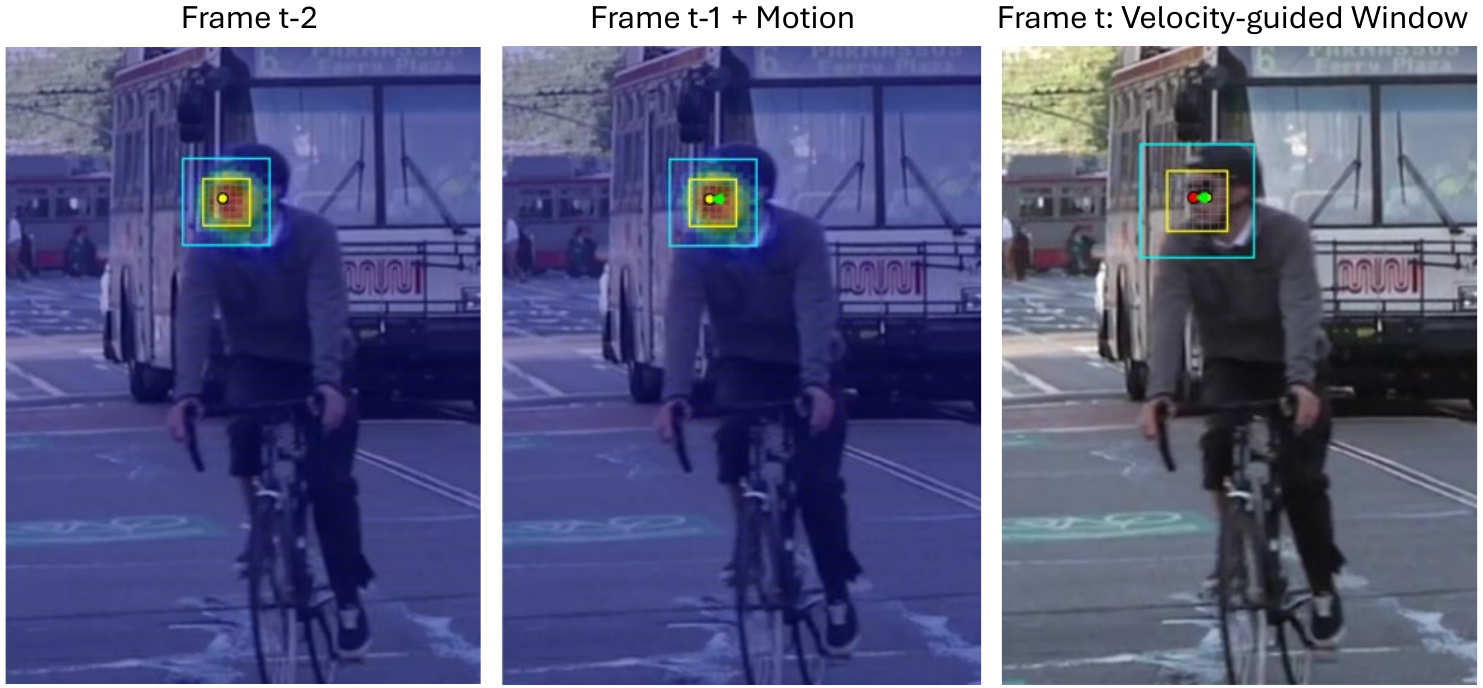}
    \caption{Illustration of the Local and Global masks for one joint. The yellow box denotes the local window, while the blue box indicates the larger global window.
    Left: past frame $t\!-\!2$ with the heatmap peak and local patch.
    Middle: past frame $t\!-\!1$ with the peak shifted by the estimated motion between $t\!-\!2$ and $t\!-\!1$.
    Right: current frame $t$, where the velocity-guided node window is centered at the extrapolated position.}
    \label{fig:node_embed}
\end{figure}

The local and global mask window sizes are $ r^{\mathrm{local}}_{t,j}=2r_{t,j}+1$,
$r^{\mathrm{global}}_{t,j}=r^{\mathrm{local}}_{t,j}+4$, as shown in~\cref{fig:node_embed}.
Binary masks \(M^{\mathrm{local}}_{t,j}\) and \(M^{\mathrm{global}}_{t,j}\) are centered at 
\((\tilde{x}^{i}_{t,j}, \tilde{y}^{i}_{t,j})\) (\cref{sec:current_joint_em}), with the window region set to $1$ and other positions to $0$.

For past frames $t-2$ and $t-1$, masks are centered at 
their Taylor-refined joint coordinates with radii $3$ and $6$ (~\cref{fig:node_embed}), capturing local refinement and broader context, respectively.

\paragraph{Node Embedding.}

Spatial attention weights are obtained via
\begin{equation}
a_{t',j,h,w}
=
\frac{\exp(\ell_{t',j,h,w})}
{\sum_{h',w'} \exp(\ell_{t',j,h',w'})},
\end{equation}

and the attended context is
$\boldsymbol{c}_{t',j}
=
\sum_{h,w}
a_{t',j,h,w}\,\boldsymbol{V}_{t',h,w}.$
The final node embedding for joint $j$ is formulated by:
\begin{equation}
\boldsymbol{z}_{t',j}
=
W_O^{\top}\boldsymbol{c}_{t',j} + \boldsymbol{b}_O
\in\mathbb{R}^{F}.
\end{equation}

Stacking embeddings across all joints and frames gives two node-embedding sets for the clip:
\[
\boldsymbol{Z}^{\mathrm{local}} \in \mathbb{R}^{T\times J\times F}, ~~~
\boldsymbol{Z}^{\mathrm{global}} \in \mathbb{R}^{T\times J\times F}.
\]

\subsection{Dual-Branch Decoupled Spatio-Temporal Attention Graphs (DSTAG)}
\label{sec:DSTAG}

Built on the joint embeddings, DSTAG performs decoupled temporal propagation followed by spatial constraint reasoning in both local and global branches.

\subsubsection{Temporal GAT for joint-wise reasoning}
\label{sec:T-GAT}

We use a temporal Graph Attention Network (GAT)~\cite{velivckovic2018graph} to model joint-wise motion across the causal window. By attending only to adjacent time steps, the module enforces causality, smooths noisy predictions, and captures short-term motion patterns.

For the three-frame window ${t-2,t-1,t}$, each joint forms a chain-structured graph
$(t-2)\rightarrow(t-1)\rightarrow t$,
where embeddings
$\bigl[\boldsymbol{z}{t-2,j},\boldsymbol{z}{t-1,j},\boldsymbol{z}_{t,j}\bigr]$
are updated via:
\begin{equation}
\mathrm{Feat}_{\mathrm{temp}}
=
\mathrm{T\text{-}GAT}(\boldsymbol{Z})
\in\mathbb{R}^{T\times J\times F_t}.
\end{equation}
We split the temporally aggregated features into a past sequence $\mathrm{Feat}_{\text{past}}
=
\mathrm{Feat}_{\text{temp}}[1{:}T-1,:,:]$ and the current-frame
representation $\boldsymbol{f}_{\text{curr}} = \mathrm{Feat}_{\text{temp}}[T,:,:].$

To maintain strict causality, we build a past-only temporal memory.
A Transformer encoder processes $\mathrm{Feat}{\text{past}}$ to summarize recent motion, and its output is fused with $\boldsymbol{f}{\text{curr}}$:
\begin{equation}
\boldsymbol{f}_{\text{temp}}
=
\mathcal{F}_{\text{fuse}}
\Bigl(
\boldsymbol{f}_{\text{curr}},
\;
\mathcal{T}_{\mathrm{enc}}
\bigl(
\mathrm{Feat}_{\text{past}}
\bigr)
\Bigr),
\end{equation}
where
$ \mathcal{T}_{\mathrm{enc}}(\cdot) \rightarrow \mathbb{R}^{J\times F_t}$ is the Transformer encoder that summarizes past history
and $\mathcal{F}_{\text{fuse}}(\cdot,\cdot) \rightarrow \mathbb{R}^{J\times F_t}$ is the adaptive fusion module. 
This produces the local and global temporally contextualized embeddings
$\boldsymbol{f}^{local}_{\text{temp}} \in \mathbb{R}^{J\times F_t}$ and $\boldsymbol{f}^{global}_{\text{temp}} \in \mathbb{R}^{J\times F_t}$.

\subsubsection{Spatial Reasoning}
After temporal reasoning, spatial relations among joints are modeled using GAT. Spatial reasoning operates on the temporally contextualized node embeddings ($\boldsymbol{f}^{local}_{\text{temp}}$ and $\boldsymbol{f}^{global}_{\text{temp}}$) at the current frame and refines them through structured message passing on the pose graph.

\begin{figure}[htbp]
    \centering
    \begin{subfigure}{0.48\linewidth}
        \centering
        \includegraphics[width=\linewidth]{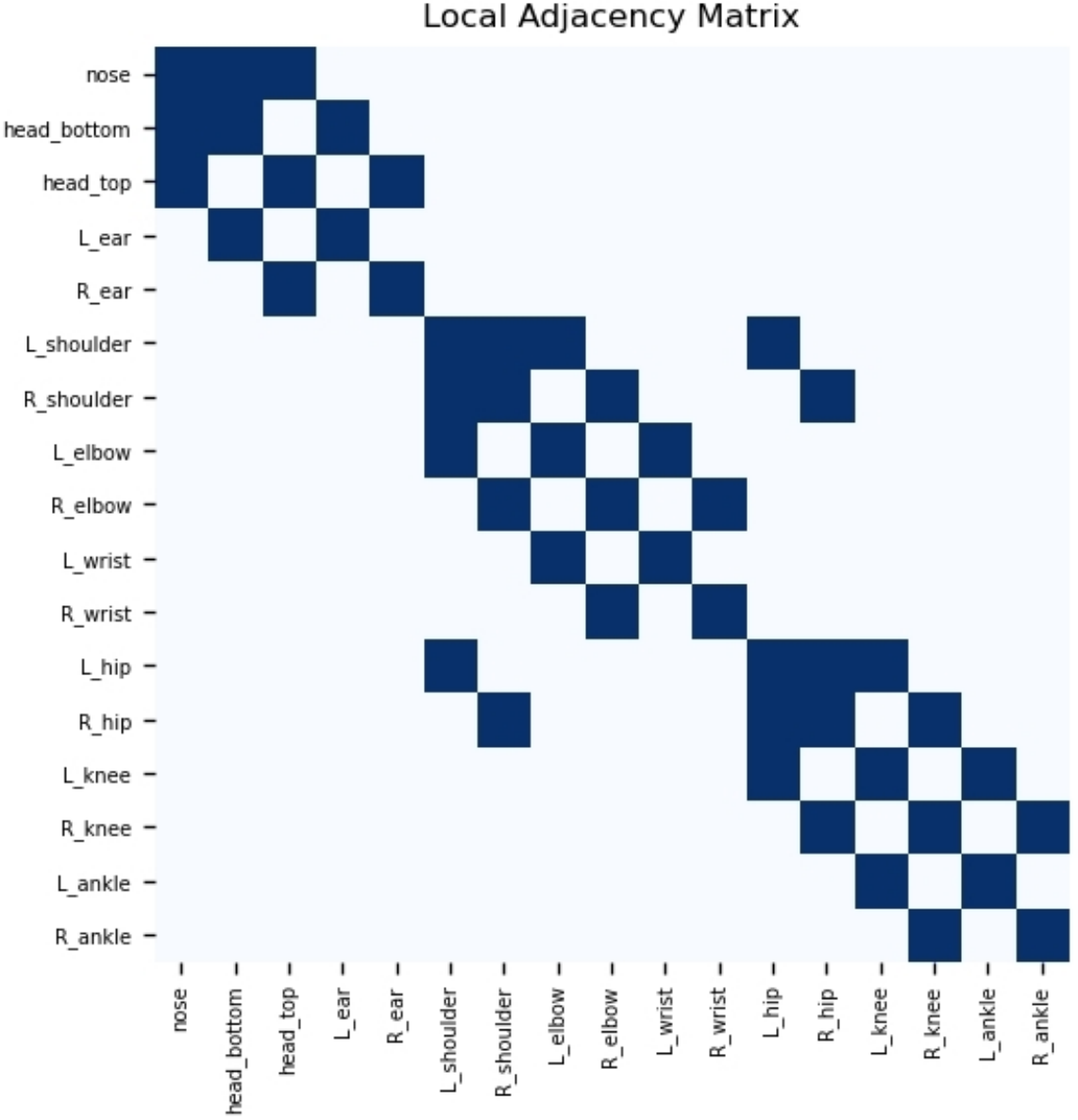}
        \caption{ }
        \label{fig:local-GAT}
    \end{subfigure}
    \hfill
    \begin{subfigure}{0.48\linewidth}
        \centering
        \includegraphics[width=\linewidth]{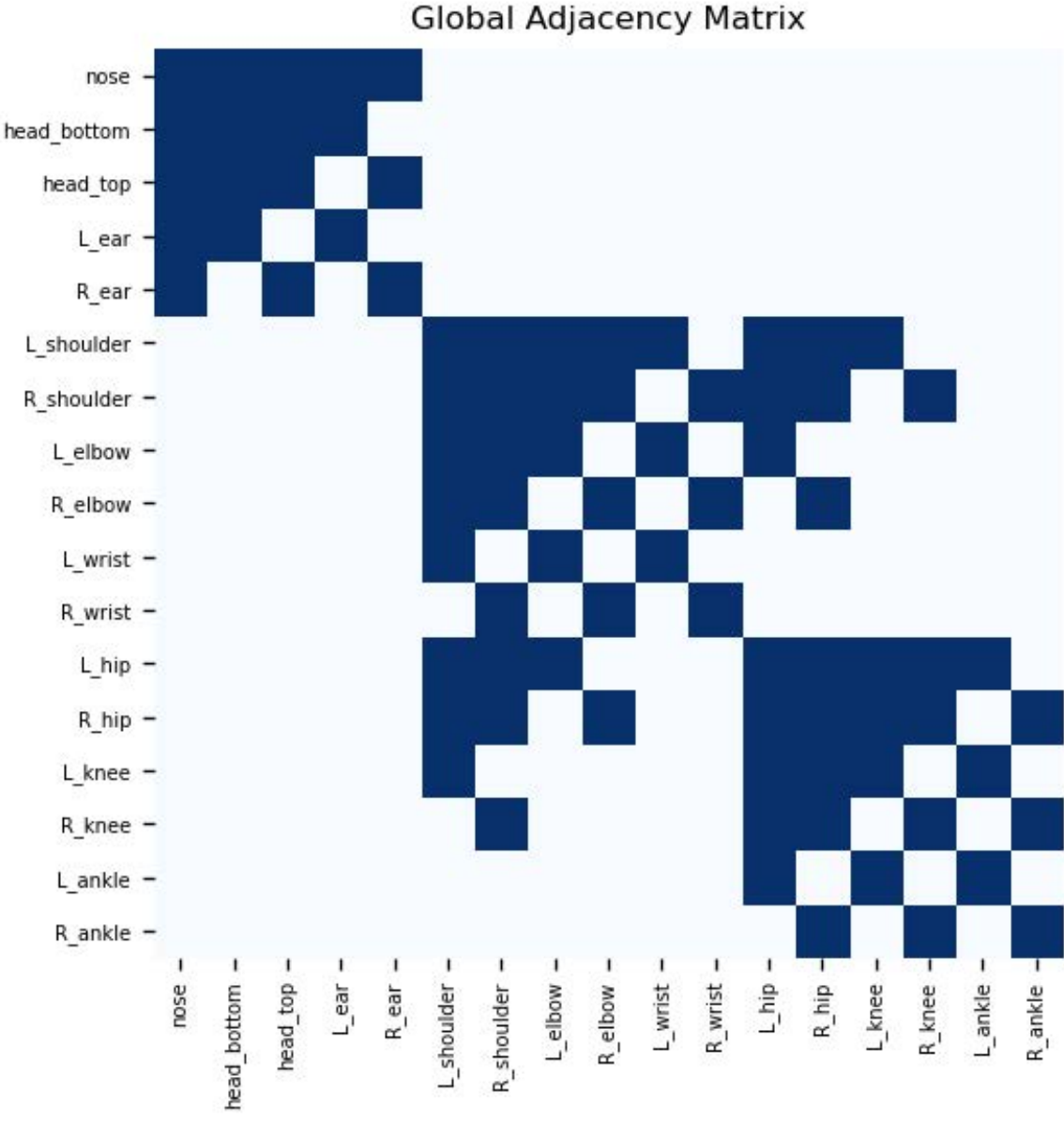}
        \caption{ }
        \label{fig:global-GAT}
    \end{subfigure}
    \caption{Adjacency matrices used in the spatial GAT branches. Each row and column corresponds to a body joint, and nonzero entries indicate neighboring joints. (a) The 1-hop skeleton adjacency matrix used in the local spatial GAT branch. (b) The 2-hop skeleton adjacency matrix used in the global spatial GAT branch.}
    \label{fig:combined}
\end{figure}

\paragraph{Local Spatial GAT (S-GAT).}
The local branch uses a fixed 1-hop skeletal adjacency matrix (\cref{fig:local-GAT}), aggregating information only from anatomically adjacent joints. This constrained message passing enforces anatomical structure and captures fine-grained part-level relations:
\begin{equation}
\boldsymbol{z}^{\text{local}}
=
\mathrm{S\text{-}GAT}\!\left(
\boldsymbol{f}^{local}_{\text{temp}},
\mathcal{A}^{\text{local}}
\right),
\end{equation}
where
$\mathcal{A}^{\text{local}} \in \{0,1\}^{J\times J}$ is the adjacency,
$\mathrm{S\text{-}GAT}(\cdot):\left(\mathbb{R}^{J\times F_t},\,\mathcal{A}^{\text{local}}\right)
\rightarrow \mathbb{R}^{J\times F_t}$ is the local spatial operator, and
$\boldsymbol{z}^{\text{local}} \in \mathbb{R}^{J\times F_t}$ is the refined local representation.

\paragraph{Global Spatial GAT.}
To capture higher-level body configuration, the global branch uses a 2-hop expanded skeleton graph (\cref{fig:global-GAT}). This enlarged receptive field allows joints to attend to more distant yet structurally relevant body parts while avoiding the noise of fully dense attention:
\begin{equation}
\boldsymbol{z}^{\text{global}}
=
\mathrm{S\text{-}GAT}\!\left(
\boldsymbol{f}^{global}_{\text{temp}},
\mathcal{A}^{\text{global}}
\right),
\end{equation}

\subsubsection{Bi-directional Cross-Branch Attention}
After local and global spatial reasoning, the two branches provide complementary cues: the local branch captures fine-grained anatomical relations, while the global branch encodes long-range body configuration. To integrate both, we apply a bi-directional cross-branch attention at the current time step, where global features attend to local features and vice versa. This reciprocal exchange refines each joint representation with information from the other branch, yielding spatially coherent and contextually enriched embeddings:
\begin{equation}
\bigl(
\boldsymbol{\tilde{z}}^{\text{local}},
\;
\boldsymbol{\tilde{z}}^{\text{global}}
\bigr)
=
\mathcal{C}\text{-}\mathrm{Attn}
\Bigl(
\boldsymbol{z}^{\text{local}},
\;
\boldsymbol{z}^{\text{global}}
\Bigr),
\end{equation}
where $\boldsymbol{\tilde{z}}^{\text{local}}\in \mathbb{R}^{J\times F_t}$
 denotes local features updated using global context, $\boldsymbol{\tilde{z}}^{\text{global}}\in \mathbb{R}^{J\times F_t}$
denotes global features updated using local cues, and $\mathcal{C}\text{-}\mathrm{Attn}(\cdot,\cdot):\left(\mathbb{R}^{J\times F_t},\,\mathbb{R}^{J\times F_t}\right)
\rightarrow \left(\mathbb{R}^{J\times F_t},\,\mathbb{R}^{J\times F_t}\right)$ denotes the bi-directional cross-branch attention operator.

\subsection{Node-Space Expert Fusion}
\label{sec:NSEF}
After dual-branch spatio-temporal reasoning, each joint obtains two complementary node predictions: a \emph{global} node that captures long-range structural context and a \emph{local} node that preserves fine-grained anatomical consistency. Rather than fusing heatmaps or intermediate logits, we combine these two experts directly in the normalized node space using a mixture-of-experts fusion. This produces a single, geometry-consistent node representation for each joint, which is subsequently rendered into a heatmap via the fixed GNC-Decoding (\cref{sec:GNC-D}).

Let $\boldsymbol{\tilde{z}}^{\text{local}}_j,\,\boldsymbol{\tilde{z}}^{\text{global}}_j \in \mathbb{R}^{F}$ 
denote the final local and global embeddings for joint $j$.  
Their node predictions are obtained by an MLP:
\begin{equation}
\begin{aligned}
\boldsymbol{u}^{(l)}_j
&=
\bigl[
x^{\text{norm}}_j,\,
y^{\text{norm}}_j,\,
v_j
\bigr]
=
f_{\mathrm{MLP}}\!\left(\boldsymbol{\tilde{z}}^{\text{local}}_j\right),
\\
\boldsymbol{u}^{(g)}_j
&=
\bigl[
x^{\text{norm}}_j,\,
y^{\text{norm}}_j,\,
v_j
\bigr]
=
f_{\mathrm{MLP}}\!\left(\boldsymbol{\tilde{z}}^{\text{global}}_j\right).
\end{aligned}
\end{equation}
To fuse the two experts, we first construct a gating input
\begin{equation}
\boldsymbol{h}_j
=
\mathrm{LayerNorm}
\!\left(
\bigl[
\boldsymbol{\tilde{z}}^{\text{local}}_j,\;
\boldsymbol{\tilde{z}}^{\text{global}}_j
\bigr]
\right)
\in \mathbb{R}^{2F}.
\end{equation}
A gating MLP predicts two scores:
\begin{equation}
\boldsymbol{s}_j
=
W_2^\top
\mathrm{ReLU}
\!\left(
W_1^\top \boldsymbol{h}_j + \boldsymbol{b}_1
\right)
+
\boldsymbol{b}_2
=
\begin{bmatrix}
\alpha \\
\beta
\end{bmatrix}.
\end{equation}
We enforce non-negativity and normalize the resulting weights:
$\alpha_j = \tilde{\alpha}_j/({\tilde{\alpha}_j + \tilde{\beta}_j})$ and
$\beta_j  = \tilde{\beta}_j/({\tilde{\alpha}_j + \tilde{\beta}_j})$,
where
$
\tilde{\alpha}_j = \mathrm{Softplus}(\alpha)
$
and $\tilde{\beta}_j  = \mathrm{Softplus}(\beta).$
The node prediction is then obtained by a convex combination:
\begin{equation}
\boldsymbol{u}^{(f)}_j
=
\alpha_j\,\boldsymbol{u}^{(g)}_j
+
\beta_j\,\boldsymbol{u}^{(l)}_j = [x^{\text{norm}}_j,y^{\text{norm}}_j,v_j].
\end{equation}
Finally, the fixed GNC-Decoding (\cref{sec:GNC-D}) converts
$\boldsymbol{u}^{(f)}_j$ into a Gaussian heatmap $\mathcal{H}_j$.

\subsection{Node-Centric Visibility-Aware Loss}
We supervise predictions directly in node space using geometric and visibility terms.
Let
$\boldsymbol{u}^{\text{pred}}_{b,j}
=[x^{\text{pred}}_{b,j},\,y^{\text{pred}}_{b,j},\,v^{\text{pred}}_{b,j}]$
and the ground truth node (\cref{sec:GNC-E}) be
$\boldsymbol{n}_{b,j}
=[x^{\ast}_{b,j},\,y^{\ast}_{b,j},\,v^{\ast}_{b,j}]$
for batch index $b$ and joint $j$.
The geometric loss applies a Huber penalty to normalized coordinates:
\begin{equation}
\mathcal{L}_{\mathrm{geo}}
=
\frac{1}{BJ}
\sum_{b=1}^{B}\sum_{j=1}^{J}
\Bigl[
g\!\left(x^{\text{pred}}_{b,j}-x^{\ast}_{b,j}\right)
+
g\!\left(y^{\text{pred}}_{b,j}-y^{\ast}_{b,j}\right)
\Bigr].
\end{equation}
Visibility is supervised with binary cross entropy:
\begin{equation}
\mathcal{L}_{\text{vis}}
=
\frac{1}{BJ}
\sum_{b=1}^{B}\sum_{j=1}^{J}
\mathrm{BCE}\!\left(
v^{\text{pred}}_{b,j},\,v^{\ast}_{b,j}
\right).
\end{equation}
The total loss is 
$\mathcal{L}
= \mathcal{L}_{\text{geo}} + \mathcal{L}_{\text{vis}}.$

\subsection{Geometric Node Codec (GNC)}
\label{sec:GNC}
GNC is a parameter-free codec providing a bidirectional mapping between geometric joint representations and heatmaps. During training, GNC-Encoding converts ground-truth heatmaps to geometric nodes. At inference, GNC-Decoding converts predicted nodes back into Gaussian heatmaps.

\subsubsection{GNC-Encoding}
\label{sec:GNC-E}
Given the ground-truth heatmap $\mathcal{H}_{t,j}\in\mathbb{R}^{H\times W}$, we extract the peak:
$(y_{t,j},\,x_{t,j})
=
\operatorname*{arg\,max}_{h,w}\,
\mathcal{H}_{t,j}.$
We then normalize coordinates:
\begin{equation}
\left\{
\begin{aligned}
\bigl[
x^{\mathrm{norm}}_{t,j},\;
y^{\mathrm{norm}}_{t,j}
\bigr]
&=
\bigl[
x_{t,j}/W,\;
y_{t,j}/H
\bigr],\\[4pt]
v_{t,j} & \in \{0,1\} \;\; 
\end{aligned}
\right.
\end{equation}

\subsubsection{GNC-Decoding}
\label{sec:GNC-D}
Given a fused node prediction $\boldsymbol{u}^{(f)}_{i,j}$, normalized coordinates are converted to heatmap indices:
$
c^{x}_{i,j}=x^{\mathrm{norm}}_{i,j}\cdot W \text{ and }
c^{y}_{i,j}=y^{\mathrm{norm}}_{i,j}\cdot H.
$
using precomputed grids
$\mathrm{xs}\in\mathbb{R}^{W}$ and
$\mathrm{ys}\in\mathbb{R}^{H}$ and a fixed Gaussian width
$\sigma>0$, the squared-distance energy is
\begin{equation}
E_{i,j}(h,w)
=
\frac{
(\mathrm{xs}[w]-c^{x}_{i,j})^2
+
(\mathrm{ys}[h]-c^{y}_{i,j})^2
}{
2\sigma^2+\varepsilon
}
\end{equation}
with $\sigma=3$ and $\varepsilon=10^{-6}$.

The deterministic heatmap is
\begin{equation}
\hat{H}_{i,j,h,w}
=
v_{i,j}\exp\bigl(-E_{i,j}(h,w)\bigr).
\end{equation}

%% file: sec/3_experiment.tex
\section{Experiments}

\subsection{Datasets and Evaluation Metrics}

Three standard VHPE benchmarks are used for performance evaluation: PoseTrack17~\cite{iqbal2017posetrack}, PoseTrack18~\cite{andriluka2018posetrack}, and PoseTrack21~\cite{doering2022posetrack21}.
PoseTrack17 provides 250 training and 50 validation clips, while PoseTrack18 expands them to 593 and 170 clips with 153{,}615 annotated poses. 
Both datasets offer 15 joints with visibility flags; training clips are sparsely labeled (middle 30 frames), and validation clips are densely annotated every four frames.
PoseTrack21 adopts the same sequences as PoseTrack18 but re-annotates them with improved boxes and additional small or occluded persons. It uses 17 COCO compatible keypoints~\cite{lin2014microsoft}, where the two ear joints serve as placeholders and are ignored during evaluation~\cite{doering2022posetrack21}.
Following the standard protocol, the mean Average Precision (mAP) is used across all three benchmarks.

\subsection{Experimental Results}
We evaluate our method on the PoseTrack17, PoseTrack18, and PoseTrack21 benchmarks using the ViTPose Huge backbone.
Tables~\ref{tab:pt21_sota}, \ref{tab:pt18_sota}, and \ref{tab:pt17_sota} report per-joint AP, overall mAP, and the temporal window adopted by each method, with the best score in \textbf{bold} and the second best \underline{underlined}.
As shown in~\cref{tab:temporal_span}, most existing approaches rely on long symmetric windows that include future frames, such as $(t-2,\dots,t+2)$ or $(t-3,\dots,t+3)$.
In contrast, our method uses a short causal window $(t-2,,t-1,,t)$ and requires no access to future frames, aligning with real-time constraints and eliminating future-frame latency.

\begin{table}[t]
\centering
\caption{
Temporal window configuration for each method. Unlike the majority of offline methods that rely on future frames, our approach is fully online and processes only past and current frames.
}
\footnotesize
{
\begin{tabular}{l c}
\toprule[1pt]
\textbf{Method} & \textbf{Frames used} \\
\midrule[1pt]
PoseWarper~\cite{bertasius2019learning} & $(t\!-\!3,\dots,t\!+\!3)$ \\
DCPose~\cite{liu2021deep}              & $(t\!-\!1,\,t,\,t\!+\!1)$ \\
SLT-Pose~\cite{fu2023improving}        & $(t\!-\!2,\dots,t\!+\!2)$ \\
FAMI-Pose~\cite{liu2022temporal}       & $(t\!-\!2,\dots,t\!+\!2)$ \\
DGNN~\cite{yang2021learning}           & $(t\!-\!3,\,t\!-\!2,\,t\!-\!1,\,t)$ \\
DSTA~\cite{he2024video}                & $(t\!-\!2,\dots,t\!+\!2)$ \\
TDMI-ST~\cite{feng2023mutual}          & $(t\!-\!2,\dots,t\!+\!2)$ \\
JM-Pose~\cite{wu2024joint}             & $(t\!-\!1,\,t,\,t\!+\!1)$ \\
DiffPose~\cite{feng2023diffpose}       & $(t\!-\!2,\dots,t\!+\!2)$ \\
TIPose~\cite{zhang2025temporal}        & $(t\!-\!2,\dots,t\!+\!2)$ \\
\textbf{Ours}                          & $(t\!-\!2,\,t\!-\!1,\,t)$ \\
\bottomrule[1pt]
\end{tabular}
}
\label{tab:temporal_span}
\end{table}

It is common in the PoseTrack literature for methods not to release the person detection outputs underlying their reported pose AP scores. Many works provide only final keypoint AP numbers without the corresponding bounding box files. To the best of our knowledge, only DSTA~\cite{he2024video}, FAMI-Pose~\cite{liu2022temporal}, PoseWarper~\cite{bertasius2019learning}, and DCPose~\cite{liu2021deep} publish their detection outputs or detector protocols. We therefore include all reported scores in Tables~\ref{tab:pt21_sota}-\ref{tab:pt17_sota}, but treat methods without released detections as reference entries, while PoseWarper, DCPose, FAMI-Pose, and DSTA are directly comparable to our YOLOv3-based evaluation pipeline.

\subsubsection{Results on PoseTrack21}
PoseTrack21 is the most recent and challenging dataset in the PoseTrack family, featuring re-annotated and tighter person boxes together with many small and heavily occluded subjects. Compared with PoseTrack17 and 18, it places greater emphasis on fine-grained joint localization, visibility reasoning, and robustness under severe appearance degradation. As shown in~\cref{tab:pt21_sota}, our method achieves an mAP of \textbf{85.1}, outperforming all prior approaches that use short-range temporal windows and surpassing several methods that rely on longer, non-causal temporal contexts.

\begin{table}[ht]
\centering
\caption{
PoseTrack21 comparison with per-joint AP and mean.
}
\scriptsize
\setlength{\tabcolsep}{1.5pt} 
\resizebox{\columnwidth}{!}{%
\begin{tabular}{l c c c c c c c c}
\toprule[1pt]
\textbf{Method} & \textbf{Head} & \textbf{Shoulder} & \textbf{Elbow} & \textbf{Wrist} & \textbf{Hip} & \textbf{Knee} & \textbf{Ankle} & \textbf{Mean} \\
\midrule[1pt]
DCPose~\cite{liu2021deep}
& 83.2 & 84.7 & 82.3 & 78.1 & 80.3 & 79.2 & 73.5 & 80.5 \\
FAMI-Pose~\cite{liu2022temporal}
& 83.3 & 85.4 & 82.9 & 78.6 & 81.3 & 80.5 & 75.3 & 81.2 \\ 
DiffPose~\cite{feng2023diffpose}
& 84.7 & 85.6 & 83.6 & 80.8 & 81.4 & \underline{83.5} & \underline{80.0} & 82.9 \\
DSTA~\cite{he2024video}
& \underline{87.5} & 87.0 & 84.2 & 81.4 & 82.3 & 82.5 & 77.7 & 83.5 \\
TDMI-ST~\cite{feng2023mutual}
& 86.8 & 87.4 & 85.1 & 81.4 & 83.8 & 82.7 & 78.0 & 83.8 \\
JM-Pose~\cite{wu2024joint}
& 85.8 & \textbf{88.1} & \underline{85.7} & \underline{82.5} & \textbf{84.1} & 83.1 & 78.5 & 84.0 \\
TIPose~\cite{zhang2025temporal}
& \textbf{87.7} & \underline{88.0} & 85.0 & 81.7 & 83.4 & 82.8 & 78.3 & \underline{84.1} \\
\midrule
\textbf{Ours}
& \textbf{87.7} & \underline{88.0} & \textbf{86.2} & \textbf{83.6} & \underline{84.0} & \textbf{84.4} & \textbf{80.3} & \textbf{85.1} \\
\bottomrule[1pt]
\end{tabular}%
}
\label{tab:pt21_sota}
\end{table}

The gains are most evident on articulation-sensitive joints such as \emph{elbow}, \emph{knee}, and \emph{ankle}, where the node-centric representation and explicit visibility modeling provide clean geometric supervision even under tight boxes and frequent occlusion. The causal temporal GAT propagates motion cues within a three-frame window, while the past-only temporal memory captures short-term structure without relying on future frames. This decoupled and fully causal temporal modeling suits PoseTrack21, where future-frame access is unrealistic and longer-range aggregation often amplifies noise from unstable detections on small subjects.

Overall, the results show that our node-centric spatio-temporal attention framework offers a robust and effective solution for VHPE, particularly under the stringent localization and occlusion conditions characteristic of PoseTrack21.

\subsubsection{Results on PoseTrack18}
As shown in~\cref{tab:pt18_sota}, our method achieves an mAP of 84.5 on PoseTrack18, outperforming most competing methods while using only a causal three-frame window. Despite relying solely on past frames, it attains top accuracy on key joints (e.g., \emph{head}, \emph{elbow}, \emph{knee}), demonstrating strong robustness to motion blur and occlusion. Compared with methods using longer or non-causal temporal windows, our node-centric spatio-temporal reasoning provides comparable or superior precision with lower temporal latency, confirming both its effectiveness and suitability for real-time deployment.

\begin{table}[ht]
\centering
\caption{
PoseTrack18 comparison with per-joint AP and mean.
}
\footnotesize
\setlength{\tabcolsep}{1.5pt}
\resizebox{\columnwidth}{!}{%
\begin{tabular}{l c c c c c c c c}
\toprule[1pt]
\textbf{Method} & \textbf{Head} & \textbf{Shoulder} & \textbf{Elbow} & \textbf{Wrist} & \textbf{Hip} & \textbf{Knee} & \textbf{Ankle} & \textbf{Mean} \\
\midrule[1pt]
PoseWarper~\cite{bertasius2019learning}
& 79.9 & 86.3 & 82.4 & 77.5 & 79.8 & 78.8 & 73.2 & 79.7 \\
DCPose~\cite{liu2021deep}
& 84.0 & 86.6 & 82.7 & 78.0 & 80.4 & 79.3 & 73.8 & 80.9 \\
SLT-Pose~\cite{fu2023improving}
& 84.3 & 87.5 & 83.5 & 78.5 & 80.9 & 80.2 & 74.4 & 81.5 \\
FAMI-Pose~\cite{liu2022temporal}
& 85.5 & 87.7 & 84.2 & 79.2 & 81.4 & 81.1 & 74.9 & 82.2 \\
DGNN~\cite{yang2021learning}
& 85.1 & 87.7 & 85.3 & 80.0 & 81.1 & 81.6 & 77.2 & 82.7 \\
DiffPose~\cite{feng2023diffpose}
& 85.0 & 87.7 & 84.3 & 81.5 & 81.4 & 82.9 & 77.6 & 83.0 \\
DSTA~\cite{he2024video}
& 85.9 & 88.8 & 85.0 & 81.1 & 81.5 & 83.0 & 77.4 & 83.4 \\
TDMI-ST~\cite{feng2023mutual}
& 86.7 & \underline{88.9} & 85.4 & 80.6 & \underline{82.4} & 82.1 & 77.6 & 83.6 \\
JM-Pose~\cite{wu2024joint}
& 86.6 & 88.7 & \underline{86.0} & \underline{81.6} & \textbf{83.3} & 83.2 & \underline{78.2} & 84.1 \\
TIPose~\cite{zhang2025temporal}
& \underline{86.9} & \textbf{89.0} & \underline{86.0} & \underline{81.6} & \textbf{83.3} & \underline{83.3} & 78.0 & \underline{84.2} \\
\midrule
\textbf{Ours}
& \textbf{87.8} & 88.6 & \textbf{86.5} & \textbf{81.7} & 81.8 & \textbf{84.1} & \textbf{79.6} & \textbf{84.5} \\
\bottomrule[1pt]
\end{tabular}%
}
\label{tab:pt18_sota}
\end{table}

\subsubsection{Results on PoseTrack17}

\cref{tab:pt17_sota} reports per-joint AP and overall mAP on PoseTrack17 compared with SOTA VHPE methods. Using only a three-frame causal window $(t-2,t-1,t)$, our method attains a competitive mAP of 86.4, matching or even surpassing methods that rely on substantially longer temporal contexts (five to seven frames).

\begin{table}[ht]
\centering
\caption{
PoseTrack17 comparison with per-joint AP and mean. 
}
\footnotesize
\setlength{\tabcolsep}{1.5pt}
\resizebox{\columnwidth}{!}{%
\begin{tabular}{l c c c c c c c c}
\toprule[1pt]
\textbf{Method} & \textbf{Head} & \textbf{Shoulder} & \textbf{Elbow} & \textbf{Wrist} & \textbf{Hip} & \textbf{Knee} & \textbf{Ankle} & \textbf{Mean} \\
\midrule[1pt]
PoseWarper~\cite{bertasius2019learning}
& 81.4 & 88.3 & 83.9 & 78.0 & 82.4 & 80.5 & 73.6 & 81.2 \\
DCPose~\cite{liu2021deep}
& 88.0 & 88.7 & 84.1 & 78.4 & 83.0 & 81.4 & 74.2 & 82.8 \\
SLT-Pose~\cite{fu2023improving}
& 88.9 & 89.7 & 85.6 & 79.5 & 84.2 & 83.1 & 75.8 & 84.2 \\
FAMI-Pose~\cite{liu2022temporal}
& 89.6 & 90.1 & 86.3 & 80.0 & 84.6 & 83.4 & 77.0 & 84.8 \\
DGNN~\cite{yang2021learning}
& \textbf{90.9} & 90.7 & 86.0 & 79.2 & 83.8 & 72.7 & 78.0 & 84.9 \\
DSTA~\cite{he2024video}
& 89.3 & 90.6 & 87.3 & 82.6 & 84.5 & 85.1 & 77.8 & 85.6 \\
TDMI-ST~\cite{feng2023mutual}
& 90.6 & 91.0 & 87.2 & 81.5 & 85.2 & 84.5 & 78.7 & 85.9 \\
JM-Pose~\cite{wu2024joint}
& \underline{90.7} & \textbf{91.6} & 87.8 & 82.1 & \textbf{85.9} & 85.3 & 79.2 & \underline{86.4} \\
DiffPose~\cite{feng2023diffpose}
& 89.0 & \underline{91.2} & 87.4 & \underline{83.5} & \underline{85.5} & \textbf{87.2} & \textbf{80.2} & \underline{86.4} \\
TIPose~\cite{zhang2025temporal}
& \underline{90.7} & 91.1 & \underline{87.9} & \textbf{83.6} & 85.3 & \underline{86.3} & 80.0 & \textbf{86.7} \\
\midrule
\textbf{Ours}
& 90.0 & 90.7 & \textbf{88.6} & \underline{83.5} & 84.3 & 85.9 & \underline{80.1} & \underline{86.4} \\
\bottomrule[1pt]
\end{tabular}%
}
\label{tab:pt17_sota}
\end{table}

\subsection{Computation Complexity}
\cref{tab:complexity} reports parameter counts and FLOPs for
different ViTPose backbones augmented with our temporal graph head.
The GNN contributes only about 6M parameters and a nearly constant compute
overhead across all variants, compared to 22-631M parameters in the backbone.
Thus the proposed head is lightweight relative to the backbone capacity and
can be attached to different ViTPose sizes to trade accuracy for compute,
without introducing a large additional budget.
\begin{table}[t]
\centering
\footnotesize
\caption{
Model complexity comparison.
}
\setlength{\tabcolsep}{3pt}
{
\begin{tabular}{lccc}
\toprule[1pt]
\textbf{Method} & \textbf{Params (M)} & \textbf{GFLOPs} & \textbf{GMACs} \\
\midrule[1pt]
{Ours (ViTPose-H)} & {643.7} & {573.3} & {285.6} \\
Ours (ViTPose-L) & 314.9 & 283.2 & 141.7 \\
Ours (ViTPose-B) & 96.2  & 94.0  & 47.1 \\
\textbf{Ours (ViTPose-S)} & \textbf{30.7}  & \textbf{38.6}  & \textbf{19.2} \\
\midrule
TIPose (ViTPose-H)~\cite{jiao2025spatiotemporal} & 666.8 & 585.0 & -- \\
DSTA (ViTPose-H)~\cite{he2024video}   & 631.0 & 123.9 & -- \\
DCPose (HRNet-W48)~\cite{liu2021deep} & 68.0  & 46.5  & -- \\
FAMI-Pose (HRNet-W48)~\cite{liu2022temporal}  & 64.5 & 183.0 & -- \\
PoseWarper (HRNet-W48)~\cite{bertasius2019learning} & 71.1 & 192.2 & -- \\
\bottomrule[1pt]
\end{tabular}
}
\label{tab:complexity}
\end{table}

\subsection{Comparison of Visual Results}
\begin{figure}[htbp]
    \centering
    \includegraphics[width=1\linewidth]{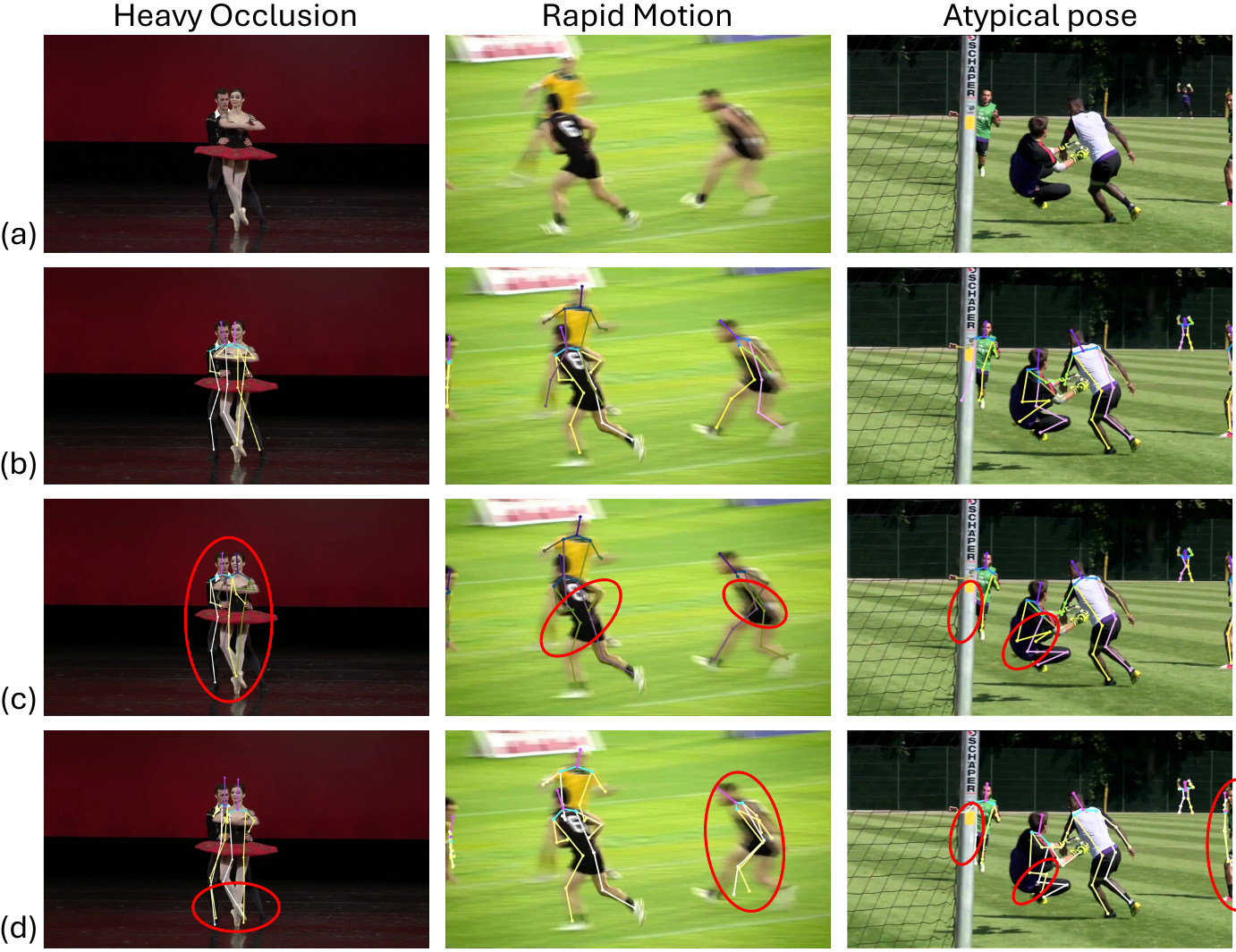}
    \caption{Comparisons on challenging scenes on PoseTrack17 including heavy occlusion, rapid motion, and atypical pose configuration. (a) Input frames; prediction from (b) Ours, (c) DSTA, and (d) DCPose. False predictions are highlighted with red circles.}
    \label{fig:Compare_Visual}
    \vspace{-0.4cm}
\end{figure}

As shown in~\cref{fig:Compare_Visual}, we assess robustness under mutual occlusion, motion blur, and atypical articulation. When people overlap, DCPose often merges nearby peaks and cross-assigns joints, while DSTA retains coarse structure but scatters distal joints and occasionally duplicates limbs. Under strong blur with partial occlusion, both baselines shorten limbs and miss lower-body joints. In irregular poses, such as the goalkeeper or players behind goal posts, they misplace joints and fail to recover hidden segments. In contrast, our method preserves limb separation through velocity-adaptive search windows, maintains structural continuity via causal past-only aggregation, and restores coherent geometry by combining global two-hop context with local adjacency constraints. The node-centric formulation focuses learning on image-conditioned joint nodes rather than heatmap texture, reducing peak collisions and improving occluded joint recovery across challenging scenarios.

\subsection{Ablation Study}
All ablations are performed on PoseTrack17, our main development benchmark. Running full ablations on PoseTrack18 and PoseTrack21 with large ViTPose backbones would be too costly and limit the number of variants we can test. We therefore use a two-stage protocol: we first assess each architectural component on PoseTrack17, then apply the final setup to PoseTrack18 and PoseTrack21. The ablated modules, including the global and local graph branches, the spatial and temporal aggregation, and the node-centric codec, are designed to be dataset agnostic. Consistent gains across all three benchmarks confirm this choice.

\subsubsection{Component Ablation}
\cref{tab:pt17_ablation_components} reports the effect of removing individual components while keeping the ViTPose Huge backbone fixed.
The full model reaches 86.4 mAP, which corresponds to a gain of 3.8 points over the single-frame ViTPose baseline in the last row.
This improvement is spread across all joints and is largest for distal keypoints such as wrists, knees, and ankles.

The rows LOCAL ONLY and GLOBAL ONLY retain only a single branch, each reducing performance by about 1.0 to 1.2 mAP. This shows that the two branches are complementary, since combining them provides a larger gain than either alone. The SPATIAL ONLY and TEMPORAL ONLY variants remove temporal or spatial reasoning entirely and perform worse than the single-branch settings. The mean AP drops by 1.5 and 1.7 points, indicating that both cues are necessary and that joint optimisation is beneficial.

The NO CENTER PREDICT variant disables the VTVJE-based center prediction and instead reuses the previous-frame center with a doubled mask size. This model still exceeds ViT Alone by 3.0 mAP but is 0.8 mAP below the full model, with most of the drop on distal joints and a slight improvement at the hip. This suggests that the center prediction provides a helpful auxiliary signal that stabilises graph refinement, especially for endpoints sensitive to small errors.

Overall, the ablations show that improvements do not stem from a single module but from the combination of global and local branches, spatial and temporal aggregation, and the auxiliary center prediction, all built on a unified node-centric representation.

\begin{table}[t]
\centering
\caption{
PoseTrack17 ablation by component (per-joint AP and mean AP). The last column reports the change in mean AP relative to the Full Model (negative means worse).
}
\scriptsize
\setlength{\tabcolsep}{1.5pt}
\resizebox{\columnwidth}{!}{%
\begin{tabular}{l c c c c c c c c c}
\toprule[1pt]
\textbf{Components} & \textbf{Head} & \textbf{Shoulder} & \textbf{Elbow} & \textbf{Wrist} & \textbf{Hip} & \textbf{Knee} & \textbf{Ankle} & \textbf{Mean} & $\Delta$\textbf{Mean}\\
\midrule[1pt]
\textbf{Full Model}
& \textbf{90.0} & \textbf{90.7} & \textbf{88.6} & \textbf{83.5} & \underline{84.3} & \textbf{85.9} & \textbf{80.1} & \textbf{86.4} & \phantom{-}0.0 \\
LOCAL ONLY
& 87.8 & \underline{90.2} & \underline{87.5} & \underline{82.7} & 83.6 & \underline{85.3} & \underline{79.2} & 85.4 & $-1.0$ \\
GLOBAL ONLY
& \underline{88.5} & \underline{90.2} & 87.0 & 82.0 & 84.1 & 85.1 & 77.7 & 85.2 & $-1.2$ \\
SPATIAL ONLY
& 87.9 & 89.7 & 87.1 & 82.1 & 83.2 & 84.6 & 78.2 & 84.9 & $-1.5$ \\
TEMPORAL ONLY
& 87.4 & 89.7 & 86.9 & 81.7 & 83.3 & 84.2 & 78.1 & 84.7 & $-1.7$ \\
NO CENTER PREDICT
& 88.4 & 90.12 & 87.4 & \underline{82.7} & \textbf{84.74} & 85.1 & 79.0 & 85.6 & $-0.8$ \\
ViT Alone
& 88.8 & 89.1 & 82.3 & 76.8 & 82.0 & 80.3 & 75.9 & 82.6 & $-3.8$ \\
\bottomrule[1pt]
\end{tabular}%
}
\label{tab:pt17_ablation_components}
\vspace{-0.4cm}
\end{table}

\subsubsection{Backbone Ablation}
\cref{tab:pt17_backbone_ablation} studies the influence of the backbone capacity under the same training protocol. As expected, performance increases with model size. ViTPose Huge reaches 86.4 mAP, ViTPose Large slightly lower at 85.9 mAP, and the Base and Small variants obtain 83.8 and 82.3 mAP. Together with the compute breakdown in ~\cref{tab:complexity}, these results show a clear accuracy-efficiency trade-off. The GNN head adds only a small and nearly constant amount of parameters and FLOPs to each backbone, yet consistently improves over the corresponding single-frame ViTPose model. This confirms that the temporal graph head can pair with backbones of different capacities depending on compute constraints.

\begin{table}[ht]
\centering
\caption{
Backbone ablation on PoseTrack17.
}
\footnotesize
\setlength{\tabcolsep}{2pt}
{%
\begin{tabular}{l c c c c c c c c}
\toprule
\textbf{Backbone} & \textbf{Head} & \textbf{Shoulder} & \textbf{Elbow} & \textbf{Wrist} & \textbf{Hip} & \textbf{Knee} & \textbf{Ankle} & \textbf{Mean} \\
\midrule
\textbf{ViT-H}
& \textbf{90.0} & \textbf{90.7} & \textbf{88.6} & \textbf{83.5} & \textbf{84.3} & \textbf{85.9} & \textbf{80.1} & \textbf{86.4} \\
\underline{ViT-L}
& \underline{89.4} & \underline{90.1} & \underline{88.1} & \underline{83.0} & \underline{84.1} & \underline{85.1} & \underline{79.4} & \underline{85.9} \\
ViT-B
& 87.5 & 88.9 & 85.8 & 79.6 & 83.2 & 82.8 & 76.6 & 83.8 \\
ViT-S
& 88.2 & 88.1 & 84.7 & 77.1 & 81.5 & 80.7 & 73.0 & 82.3 \\
\bottomrule
\end{tabular}%
}
\label{tab:pt17_backbone_ablation}
\vspace{-0.4cm}
\end{table}

\section{Conclusion}
We presented a node-centric framework for video-based human pose estimation that explicitly unifies visual, temporal, and structural reasoning. Instead of depending on heatmaps or implicit temporal aggregation, our method constructs expressive joint representations through a visuo-temporal velocity embedding and an attention-based pose-query encoder that maps appearance and motion cues into an image-conditioned node space. A dual-branch decoupled spatio-temporal graph then performs complementary local and global reasoning, and a node-space expert fusion module adaptively integrates both to produce accurate and stable joint predictions. Experiments on three benchmarks show that our approach consistently surpasses state-of-the-art methods, highlighting the effectiveness of explicit node-centric modeling and its potential for advancing robust, fine-grained video pose estimation.